\documentclass{ifacconf}

\usepackage{graphicx}      
\usepackage{natbib}        

\usepackage{algorithmic}
\usepackage{amsmath}
\usepackage{amsfonts}
\usepackage{float}
\usepackage{savesym}
\savesymbol{AND}
\begin{document}
\begin{frontmatter}

\title{Inductive Geometric Matrix Midranges\thanksref{footnoteinfo}} 

\thanks[footnoteinfo]{This work has received support from the European Research Council under the Advanced ERC Grant Agreement Switchlet n.670645. Graham Van Goffrier acknowledges support from the Cambridge Mathematics Placement (CMP) Programme.
Cyrus Mostajeran is supported by the Cambridge Philosophical Society.}

\author[First]{Graham W. Van Goffrier} 
\author[Second]{Cyrus Mostajeran} 
\author[Second]{Rodolphe Sepulchre}

\address[First]{Department of Physics and Astronomy, University College London, London, United Kingdom (e-mail: vangoffrier@gmail.com)}
\address[Second]{Department of Engineering,
University of Cambridge, 
Cambridge CB2 1PZ,
United Kingdom (e-mail: csm54@cam.ac.uk)}

\begin{abstract}                
Covariance data as represented by symmetric positive definite 
(SPD) matrices are ubiquitous throughout technical study as efficient
descriptors of interdependent systems. Euclidean analysis
of SPD matrices, while computationally fast, can lead to skewed and even
unphysical interpretations of data. Riemannian methods preserve the
geometric structure of SPD data at the cost of expensive eigenvalue
computations. In this paper, we propose a geometric method for
unsupervised clustering of SPD data based on the Thompson metric. This technique relies upon a novel ``inductive midrange" centroid computation for SPD data, whose properties are examined and numerically confirmed. We demonstrate the incorporation
of the Thompson metric and inductive midrange into $X$-means and $K$-means++
clustering algorithms. 
\end{abstract}


\end{frontmatter}

\section{Introduction}  \label{intro}

The arithmetic midrange of a finite collection of real numbers $\{y_i\}_{i\in I}$ is defined as the mean of the extremal values: $\frac{1}{2}(\min_i y_i + \max_i y_i)$. This number can also be uniquely characterized as the solution $x^*$ to the optimization problem
\begin{equation}
\min_x \max_i|x-y_i|.
\end{equation}
One can also characterize the midrange as the limit of an inductive procedure on the input data points $y_i$ in the following way. The merits of such a characterization will become clear in due course. 

\begin{prop} \label{inductive scalars}
For any $a,b\in \mathbb{R}$, define the curve $\gamma(a,b;\cdot):[0,1]\to \mathbb{R}$ by $\gamma(a,b;t)=(1-t)a+tb$. For any initialization $x_1\in\mathbb{R}$, we may generate a sequence $(x_k)$ as follows. Given a point $x_k$:
\begin{enumerate}
\item choose a point $y_k^{\uparrow}\in\{y_i\}$ such that $|x_k-y_k^{\uparrow}| \geq |x_k-y_i|$ for all $i\in I$; 
\item set $x_{k+1}:=\gamma(x_k,y_k^{\uparrow};\frac{1}{k+1})$.
\end{enumerate}
Any sequence $(x_k)$ generated as above converges to the midrange of $\{y_i\}$.
\end{prop}

\begin{pf}
Denote by $y_{\min}$ and $y_{\max}$ the minimum and maximum values within $\{y_i\}$ and write $x^*=\frac{1}{2}(y_{\min}+y_{\max})$.
For any sequence $(x_k)$ generated as above, there exists $m\in\mathbb{N}$ such that $x_m\in(y_{\min},x^*)$ and $x_{m+1}\in (x^*,y_{\max})$. Such an $m$ can easily be found by taking a sufficiently large $m$. We then have
\begin{align*}
x_{m+1}&=\frac{m}{m+1}x_m + \frac{1}{m+1}y_{\max} \\
x_{m+2} &= \frac{m+1}{m+2}x_{m+1} + \frac{1}{m+2}y_{\min},
\end{align*}
which combine to give
\begin{align*}
x_{m+2}&=\frac{m}{m+2}x_{m}+\frac{1}{m+2}(y_{\min}+y_{\max}) \\
&= \frac{m}{m+2}x_m + \frac{2}{m+2}x^*.
\end{align*}
Thus, 
\begin{align*}
x_{m+2} - x_{m} &= \frac{2}{m+2}(x^*-x_m) > 0 \\
x_{m+2} - x^* &= \frac{m}{m+2}(x_m-x^*) < 0,
\end{align*}
so that $x_m < x_{m+2} < x^*$. By symmetry in the reasoning, we also have $x^* < x_{m+3} < x_{m+1}$. By induction on $k$, we obtain
\begin{equation}
x_{m+2k}-x^*=\frac{m}{m+2k}(x_m-x^*) \leq \frac{m}{2k}(x_m-x^*)
\end{equation}
for all $k\geq 1$. Thus, we see that the subsequence $(x_{m+2k})$ converges to $x^*$ at a rate of $\mathcal{O}(k^{-1})$. Since a similar result holds for the subsequence $(x_{m+2k-1})$, we conclude that $(x_k)$ converges to the midrange $x^*$ at the given rate.
\qed
\end{pf}

Although the midrange is clearly sensitive to outliers, it can be a useful measure in some circumstances. In particular, if the data points $y_i$ are uniformly spread within a domain, the mean coincides with the midrange and can thus be computed using only extremal values without sifting through all the points. Similarly, the midrange can be useful for analyzing data that is devoid of outliers and finds applications in clustering algorithms that rely on the isolation of outlying clusters (\cite{Steinley2006,k-midranges,stigler2016}).

In this paper, we consider midrange statistics in the cone of symmetric positive definite (SPD) matrices. Data representations based on SPD matrices arise in an enormous range of applications as covariance matrices, including brain-computer interface (BCI) systems (\cite{Rao2013,Salem2018}), radar data processing (\cite{Arnaudon2013}), and diffusion tensor imaging (DTI) (\cite{Dryden2009}). Let $\mathbb{P}_d$ denote the space of $d\times d$ real SPD matrices. The conic geometry of $\mathbb{P}_d$ often renders conventional Euclidean approaches to statistics and analysis on $\mathbb{P}_d$ ineffective. Indeed it is well-documented that using Euclidean algorithms on nonlinear spaces such as $\mathbb{P}_d$ often results in poor accuracy and undesirable effects, such as swelling phenomena in DTI (\cite{Log-Euclidean2006}).

Several non-Euclidean geometries have been associated with $\mathbb{P}_d$ and successfully exploited in various applications. Geometries that have been studied in great detail include the log-Euclidean geometry and the affine-invariant Riemannian geometry. The log-Euclidean geometry is derived from the Lie group structure of $\mathbb{P}_d$ under the group operation $\Sigma_1\circ \Sigma_2=\exp(\log(\Sigma_1)+\log(\Sigma_2))$ for $\Sigma_1,\Sigma_2\in\mathbb{P}_d$, where $\exp$ and $\log$ denote the usual matrix exponential and logarithm. The affine-invariant Riemannian geometry on the other hand is induced by the Riemannian metric 
\begin{equation} \label{inner product}
\langle V,W \rangle_{\Sigma}=\operatorname{tr}(\Sigma^{-1}V\Sigma^{-1}W),
\end{equation}
where $V$ and $W$ are vectors belonging to the tangent space $T_{\Sigma}\mathbb{P}_d$ at $\Sigma\in\mathbb{P}_d$. This smooth metric structure induces a well-defined distance function $d_2:\mathbb{P}_d\times\mathbb{P}_d\to [0,\infty)$ given by
\begin{equation*} 
d_2(\Sigma_1,\Sigma_2)=\left[\operatorname{tr}\left(\log^2(\Sigma_1^{-\frac{1}{2}}\Sigma_2\Sigma_1^{-\frac{1}{2}}\right)\right]^{\frac{1}{2}}=\left(\sum_{i=1}^d\log^2\lambda_i\right)^{\frac{1}{2}}
\end{equation*}
where $\lambda_i$ denote the eigenvalues of $\Sigma_2\Sigma_1^{-1}$. Moreover, the unique length-minimizing geodesic from $\Sigma_1$ to $\Sigma_2$ takes the form of the curve $\gamma(t)=\Sigma_1\#_t\Sigma_2$, where
\begin{equation} \label{R geodesic}
\Sigma_1\#_t\Sigma_2 = \Sigma_1^{1/2}\left(\Sigma_1^{-1/2}\Sigma_2\Sigma_1^{-1/2}\right)^t \Sigma_1^{1/2},
\end{equation}
for $t\in[0,1]$ (\cite{Bhatia2003}). The midpoint of this geodesic defines the geometric mean $\Sigma_1\#_{\frac{1}{2}}\Sigma_2$ of $\Sigma_1$ and $\Sigma_2$. The geometric mean of $N$ SPD matrices $Y_1,\dots,Y_N$ can be defined as the unique solution to
\begin{equation}
\operatorname{argmin}_{X\in\mathbb{P}_d}\sum_{i=1}^Nd_2(X,Y_i)^2,
\end{equation}
which is also known as the Karcher mean (\cite{Moakher2005}).

An important property of the Riemannian geometry defined by (\ref{inner product}) is affine-invariance, whereby congruence transformations form isometries of $\mathbb{P}_d$. In particular, for any matrix $A$ in the general linear group $GL(d)$, we have
$d_2(A\Sigma_1A^T,A\Sigma_2A^T)=d_2(\Sigma_1,\Sigma_2)$ for all $\Sigma_1,\Sigma_2\in\mathbb{P}_d$. Affine-invariance of algorithms at the level of SPD matrices corresponds to invariance under affine transformations of the underlying feature vectors, which is often a desirable property in applications involving covariance matrices. 

A non-Riemannian affine-invariant geometry associated with $\mathbb{P}_d$ is that induced by the Thompson metric on the cone of $d\times d$ positive semidefinite matrices (\cite{Thompson1963,Lemmens2012}). The Thompson metric $d_{\infty}$ on $\mathbb{P}_d$ takes the form
\begin{align} \label{Thompson metric}
d_{\infty}(\Sigma_1,\Sigma_2)=\log\max\{\lambda_{\max}(\Sigma_1\Sigma_2^{-1}),\lambda_{\max}(\Sigma_2\Sigma_1^{-1})\},
\end{align}
where $\lambda_{\max}(\Sigma)$ denotes the maximum eigenvalue of $\Sigma$.
Note that the Thompson metric can be rewritten as $d_{\infty}(\Sigma_1,\Sigma_2)=\max_{1\leq i \leq d}|\log \lambda_i|$, which justifies the notation $d_{\infty}$. The pair $(\mathbb{P}_d,d_{\infty})$ constitutes a complete metric space of non-positive curvature (\cite{Bhatia2003}).
While the Riemannian geodesic (\ref{R geodesic}) is also a length-minimizing geodesic of the Thompson metric, it is known that $d_{\infty}$ generally admits infinitely many geodesics between a pair of points (\cite{Nussbaum1994}). In particular, the curve $\gamma_{\infty}:[0,1]\to\mathbb{P}_d$, $\gamma_{\infty}(t)=\Sigma_1*_t\Sigma_2$ defined below is a geodesic of $d_{\infty}$ from $\Sigma_1$ to $\Sigma_2$, which satisfies a number of attractive properties. Let $\lambda_{M}$ and $\lambda_{m}$ denote the maximum and minimum eigenvalues of $\Sigma_2\Sigma_1^{-1}$. If $\lambda_{M}\neq\lambda_{m}$, we define
\begin{equation} \label{Nussbaum}
\Sigma_1*_t\Sigma_2=
\left(\frac{\lambda_{M}^t-\lambda_{m}^t}{\lambda_{M}-\lambda_{m}}\right)\Sigma_2+\left(\frac{\lambda_{M}\lambda_{m}^t-\lambda_{m}\lambda_{M}^t}{\lambda_{M}-\lambda_{m}}\right)\Sigma_1,
\end{equation}
and $\Sigma_1*_t\Sigma_2=\lambda_m^t\Sigma_1$ otherwise. One of the attractive properties of (\ref{Nussbaum}) is that its midpoint $\Sigma_1*_{1/2}\Sigma_2$ scales geometrically: 
\begin{equation*}
(a_1\Sigma_1)*_{1/2}(a_2\Sigma_2)=\sqrt{a_1a_2}(\Sigma_1*_{1/2}\Sigma_2), \quad a_1,a_2>0,
\end{equation*}
and coincides with the geometric mean when $d=2$. 

Computationally, the Thompson geodesic $\Sigma_1*_t\Sigma_2$ (\ref{Nussbaum}) is considerably less expensive to construct than the Riemannian geodesic $\Sigma_1\#_t\Sigma_2$ (\ref{R geodesic}), particularly for high dimensional matrices. This is a consequence of the fact that $\Sigma_1*_t\Sigma_2$ only relies on the computation of extremal generalized eigenvalues of the pair $(\Sigma_1,\Sigma_2)$, which can be computed efficiently by several algorithms (\cite{Golub2000,Stewart2002,Mishra2016}). Similarly, the computation of the Thompson distance between a pair of SPD matrices only relies on the evaluation of extremal generalized eigenvalues, whereas the Riemannian distance involves the full generalized eigenspectrum of the pair of points.

In the case of positive scalars $(d=1)$, affine-invariance simply reduces to invariance under  scaling in the cone of positive real numbers. The affine-invariant midrange of a collection of $N$ positive scalars $y_i$ is then the geometric mean of the extremal values, which can be formulated as the unique solution of the optimization problem
\begin{equation}
\min_{x>0}\max_{1\leq i \leq N}\bigg|\log\frac{x}{y_i}\bigg|.
\end{equation}
The affine-invariant SPD matrix analogue of the above optimization formulation of the geometric midrange takes the form
\begin{align}  \label{matrix midrange}
\min_{X \in\mathbb{P}_d} \; \max_{1\leq i \leq N} \; \|\log(Y_i^{-1/2}XY_i^{-1/2})\|_{\infty},
\end{align}
which can be recast as the convex optimization problem
\begin{equation} \label{CVX}
\begin{cases}
\min_{\tau,\xi,X\succeq 0}\xi \\
\tau Y_i \preceq X \preceq \xi Y_i \\
1/\xi - \tau \leq 0
\end{cases}
\end{equation}
where $\preceq$ denotes the matrix L{\"o}wner order (\cite{Mostajeran2018,Mostajeran2020}).  

Although (\ref{CVX}) can be solved using standard convex optimization packages, the problem does not scale well with the dimension $d$ or the number of data points $N$. Thus, the aim of this work is to develop an alternative route to a notion of a geometric midrange of SPD matrices that has favorable computational properties that scale with the dimension $d$. It is in this context that we investigate the natural generalization of the inductive procedure for computing the midrange of scalars outlined in Proposition \ref{inductive scalars} to the matrix setting as an alternative definition of the geometric midrange of a collection of points in $\mathbb{P}_d$. By using the Thompson metric $d_{\infty}$ to compute the distances in the first condition of Proposition \ref{inductive scalars}, and the Thompson geodesics $\gamma_{\infty}$ as the interpolating curves in the second condition, we arrive at an algorithm that relies only on the computation of generalized extremal eigenvalues. This algorithm and its computational properties are investigated in the following sections.

\section{Inductive Midrange}

The inductive midrange (IMR) algorithm takes as input a set of symmetric positive definite (SPD) matrix data $\{Y_i\}_{i\in I}$, as well as an initialization point which may or may not be a member of the data set, and iterates towards a representative midrange centroid of the data set. Given an initial point $X_1$, a sequence $(X_k)$ is generated according to the following process.
\begin{enumerate}
\item Choose a point $Y_k^{\uparrow}\in\{Y_i\}$ such that $d_{\infty}(X_k,Y_k^{\uparrow}) \geq d_{\infty}(X_k,Y_i)$ for all $i\in I$.
\item Set $X_{k+1}:=X_k*_{\frac{1}{k+1}}Y_k^{\uparrow}$.
\end{enumerate}

In pseudocode, the algorithm can be represented as: 
\begin{algorithmic}
\STATE $midrange[1] \gets init\_midrange$
\FORALL {$1\leq i\leq numiters $}
	\FORALL {$1\leq j\leq N $}
		\STATE $dist[j] \gets d_\infty(midrange[i],data[j])$
	\ENDFOR
	\STATE $indmax \gets maxindex(dist)$ 
	\STATE $w \gets \frac{1}{1+i}$
	\STATE $midrange[i+1] \gets M(midrange[i],data[indmax],w)$
\STATE $i\gets i+1$
\ENDFOR
\end{algorithmic}
where $M(midrange[i],data[indmax],w)$ is the weighted geometric midrange

\begin{equation} \label{eq:mr2}
M(A,B,w) = \frac{(\lambda_M^w-\lambda_m^w)B + (\lambda_M \lambda_m^w - \lambda_m \lambda_M^w)A}{\lambda_M-\lambda_m}
\end{equation}

with $\lambda_{M} = \lambda_{\max}(BA^{-1})$ and $\lambda_{m} = \lambda_{\min}(BA^{-1})$.  As shown in \cite{Lim2013}, the point $X=M(A,B,1/2)$ is a Thompson metric midpoint of $A$ and $B$. That is,
\begin{equation*}
d_{\infty}(A,X)=d_{\infty}(X,B)=\frac{1}{2}d_{\infty}(A,B).
\end{equation*}

The effect of the weighted midrange in the IMR algorithm is to produce step sizes that decrease as $\frac{1}{k}$. As outlined above, these steps are restricted to the direction of the furthest data point from each successive IMR candidate, reflecting the original intent of midrange statistics to primarily account for data outliers.

We emphasize that the convergence point of the IMR \emph{is not equivalent} to the optimization midrange discussed in Section 1. As a demonstrative example, we evaluate both midranges on the data set:

\begin{equation*}
\{Y_i\} =  \left\{\begin{pmatrix} 0.95 & -0.6 \\ -0.6 & 1.1 \end{pmatrix}, \begin{pmatrix} 1.0 &  0.5 \\  0.5 & 2.1 \end{pmatrix}, 
\begin{pmatrix} 2.5 & -0.2 \\ -0.2 & 1.2 \end{pmatrix}\right\}.
\end{equation*}

The optimization midrange ($M_{OPT}$) and IMR midrange ($M_{IMR}$) are evaluated to be:

\begin{equation*}
M_{OPT} =  \begin{pmatrix} 1.32 & -0.53 \\ -0.53 & 1.62 \end{pmatrix}, \hspace{8pt}
M_{IMR} =  \begin{pmatrix} 1.14 & -0.25 \\ -0.25 & 1.25 \end{pmatrix}.
\end{equation*}

While the $M_{OPT}$ midrange attains the minimum cost-function value of 0.790, the IMR has a nearby cost-function evaluation of 0.811, which represents a less than $3\%$ increase. The Thompson distance between these two midrange matrices is 0.33, which is modest in comparison to the $\sim1.4$ average separation between data set matrices. Thus, we observe an example of the interesting phenomenon that equivalent characterizations of a mathematical object in a linear space can generalize to distinct notions in nonlinear spaces.

The primary observed features of the IMR algorithm are a universal $\frac{1}{k}$ convergence rate, initialization-invariance, and dependence exclusively on a subset of the input data called \emph{active data}. We will demonstrate each of these features in turn through numerical studies on a broad range of data set sizes $N$ and matrix dimensions $d$.

\subsection{Numerical Results}

\begin{claim}
The IMR  algorithm converges at a rate of $\frac{1}{k}$ regardless of the size or matrix dimensionality of the data set.
\end{claim}

We observed that the IMR algorithm converges in all of our numerical experiments. The convergence rate of the IMR algorithm is assessed by measuring the Thompson distance between the IMR estimate at each step and the final IMR value after $numiters$ steps. Although we do not present an analytic expression for the IMR convergence point in this paper, we take $midrange[numiters]$ as an acceptable estimate to the true convergence point for the purposes of establishing a convergence rate.

\begin{figure*}[h]
\centering
  \includegraphics[width=0.85\linewidth]{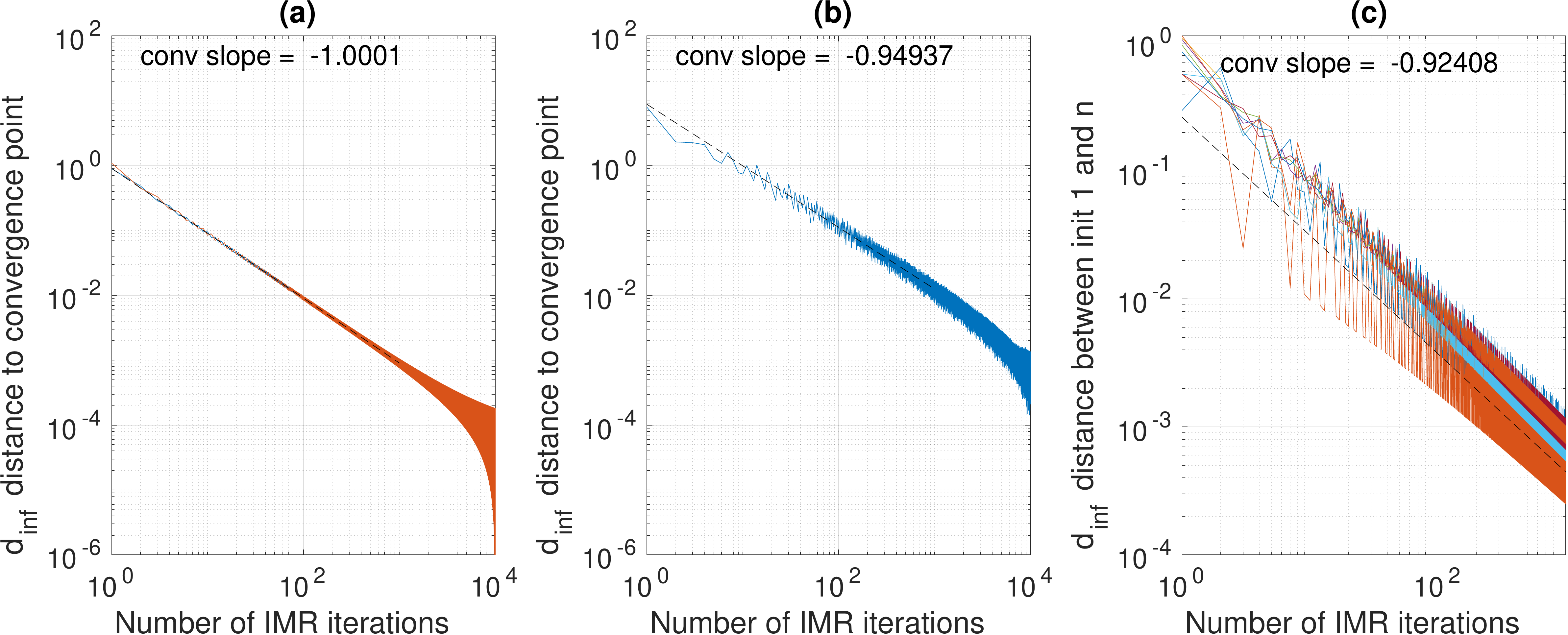}
  \caption{IMR convergence plot for (a) $d = 2$, $N = 5$ and (b) $d = 100$, $N = 5$. (c) Convergence plot showing the evolution of the distances between IMR trajectories with $d = 2$ and $N = 10$. The Thompson distances are computed from an arbitrary reference trajectory. The plot shows that all trajectories contract towards each other at approximately the specified rate.} 
  \label{fig:convplots}
\end{figure*}

Fig.~\ref{fig:convplots} (a) shows a typical example of the convergence measure throughout an IMR run for $d = 2$,  $N = 5$, where $N$ is the number of data points in $\mathbb{P}_d$. After $~10^3$ iterations, the error inherent in our estimate of the true convergence point leads to nonlinearity in this measure. Therefore the range $10^3$ to $10^4$ of iterations is excluded from fitting, and an averaged trend of $\sim\frac{1}{k}$ convergence is demonstrated. Different colors in the plot correspond to different input data taken to initialize the algorithm.
Fig.~\ref{fig:convplots} (b) shows a typical example of the same kind for $d = 100$,  $N = 5$, with only one initialization included.

Table~\ref{tb:conv} aggregates observed convergence rates for setups with larger input data sets or greater matrix dimensionality. Average convergence rates for 10 runs ($numiters = 10^4$) are given, with the fit excluding nonlinearity past $10^3$ iterations. Random SPD data are generated via the transpose-product method, as are all other SPD data in this paper unless otherwise specified.

\begin{table}[hb]
\begin{center}
\caption{Average convergence rates for several sample SPD data set configurations}\label{tb:conv}
\begin{tabular}{ccccc}
$(d,N)$ & (5,5) & (5,20) & (50,5) & (50,20) \\\hline
Rate & -0.9942 & -0.9932  & -0.9965 & -1.0019\\ \hline
\end{tabular}
\end{center}
\end{table}

\medskip

\begin{claim}
	For a given data set, the IMR algorithm converges to the same SPD matrix regardless of initialization.
\end{claim}

The input data and IMR trajectories of $2\times 2$ SPD matrices can be visualized in a cone in $\mathbb{R}^3$ as described by \cite{Mostajeran2018}, according to the bijection:

\begin{equation}
	\begin{pmatrix} a & b \\ b & c \end{pmatrix} \mapsto \left( \sqrt{2}b, \frac{1}{\sqrt{2}}(a-c), \frac{1}{\sqrt{2}}(a+c)\right).
\end{equation}

\begin{figure}[h]
\begin{center}
\includegraphics[width=0.6\linewidth]{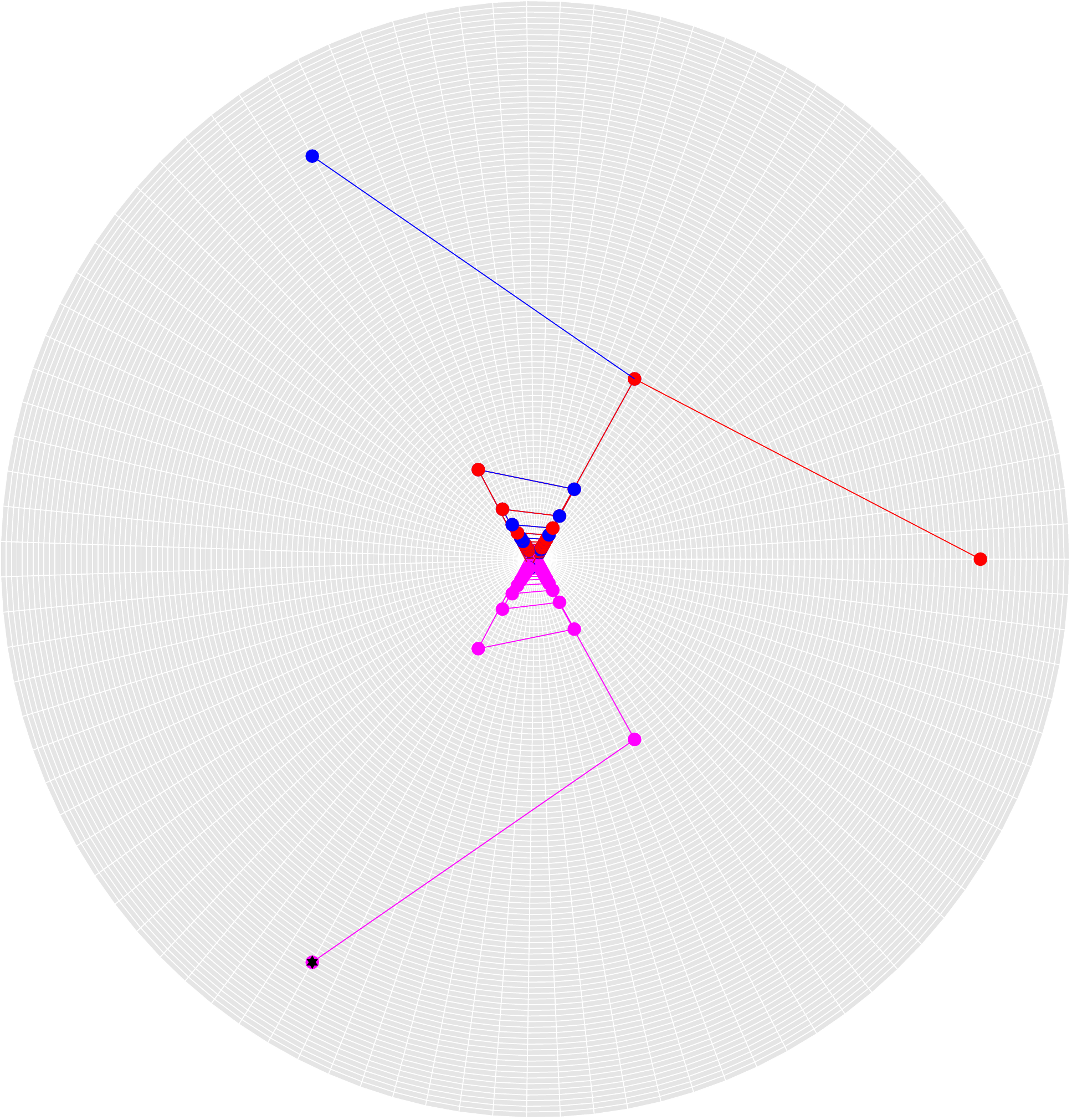}    
\caption{Invariance of the point of convergence of the IMR algorithm with respect to the choice of initialization. The plot depicts an example with $d = 2$ and $N = 3$.} 
\label{fig:cone_d2n3}
\end{center}
\end{figure}

Fig.~\ref{fig:cone_d2n3} shows a successful IMR run for $d = 2$ and  $N = 3$ depicted in a 2D projection of the cone in $\mathbb{R}^3$. Although the figure depicts a 2D projection for simplicity, the convergence is observed in the full three-dimensional space.

 The top and right-hand input data points move towards each other to their mutual midpoint during the first IMR step, and therefore trivially converge to a single result; however, it is less trivial that the bottom input data point converges to the same result.

In Fig.~\ref{fig:convplots} (c) the pairwise Thompson distances between an arbitrary reference initialization and all others are plotted throughout a $d = 2$,  $N = 10$ IMR run. Since different initializations lead to IMR paths which are drawn to some data points more often than others, quasiperiodic features dominate the pairwise distances. Despite this, all pairwise distances appear to converge to zero with a $\sim\frac{1}{k}$ rate.

To further demonstrate invariance beyond only initializations from the input data set, IMR runs were performed with randomly generated SPDs as initializations for several data set configurations. In all cases, a maximum Thompson distance separation could be identified within which all IMR results could be bounded for the chosen $numiters$ and initializations. By increasing $numiters$, the maximum separation could be reduced arbitrarily. The average separation between the trajectories after $numiters = 10^4$ steps was significantly smaller than the maximum separation in all cases, as shown in Table~\ref{tb:init}.

\begin{table}[hb]
\begin{center}
\caption{Initialization invariance results for the IMR algorithm with random initializations.}\label{tb:init}
\begin{tabular}{ccccc}
$(d,N)$ & (2,5) & (5,5) & (20,5) & (100,5) \\\hline
Max separation & .0018 & .0310 & .0864 & .1453 \\
Average separation & .0008 & .0072 & .0188 & .0277 \\
Convergence count (/100) & 100 & 100 & 100 & 100 \\ \hline
\end{tabular}
\end{center}
\end{table}

\medskip

\begin{claim}
	The IMR convergence point depends exclusively on a subset of the initial data called the \emph{active data}.  
\end{claim}

This property is essential for the IMR to be interpreted as a midrange-type centroid for a data set. For a data set $D \subset \mathbb{P}_d^N$, we define the ``external" data $E \subset D$ as the subset of all data points $D_i\in D$ that for some choice of initialization $X_1$ and $k\geq1$, 
\begin{equation}
	X_{k+1}=X_k*_{\frac{1}{k+1}}D_i.
\end{equation}
Only the external data $E$ are available for the IMR algorithm to target and step towards on each iteration. As the size of an SPD data set increases, it becomes more difficult to arrange them in such a way that $|E|$ increases.

However, simulations demonstrate that the IMR need not be dependent on all data in $E$ in the limit of infinite time, and in fact is generally dependent on a subset $A \subseteq E$ which may be significantly smaller -- we term this subset the active data.

\begin{figure}
\begin{center}
\includegraphics[width=0.6\linewidth]{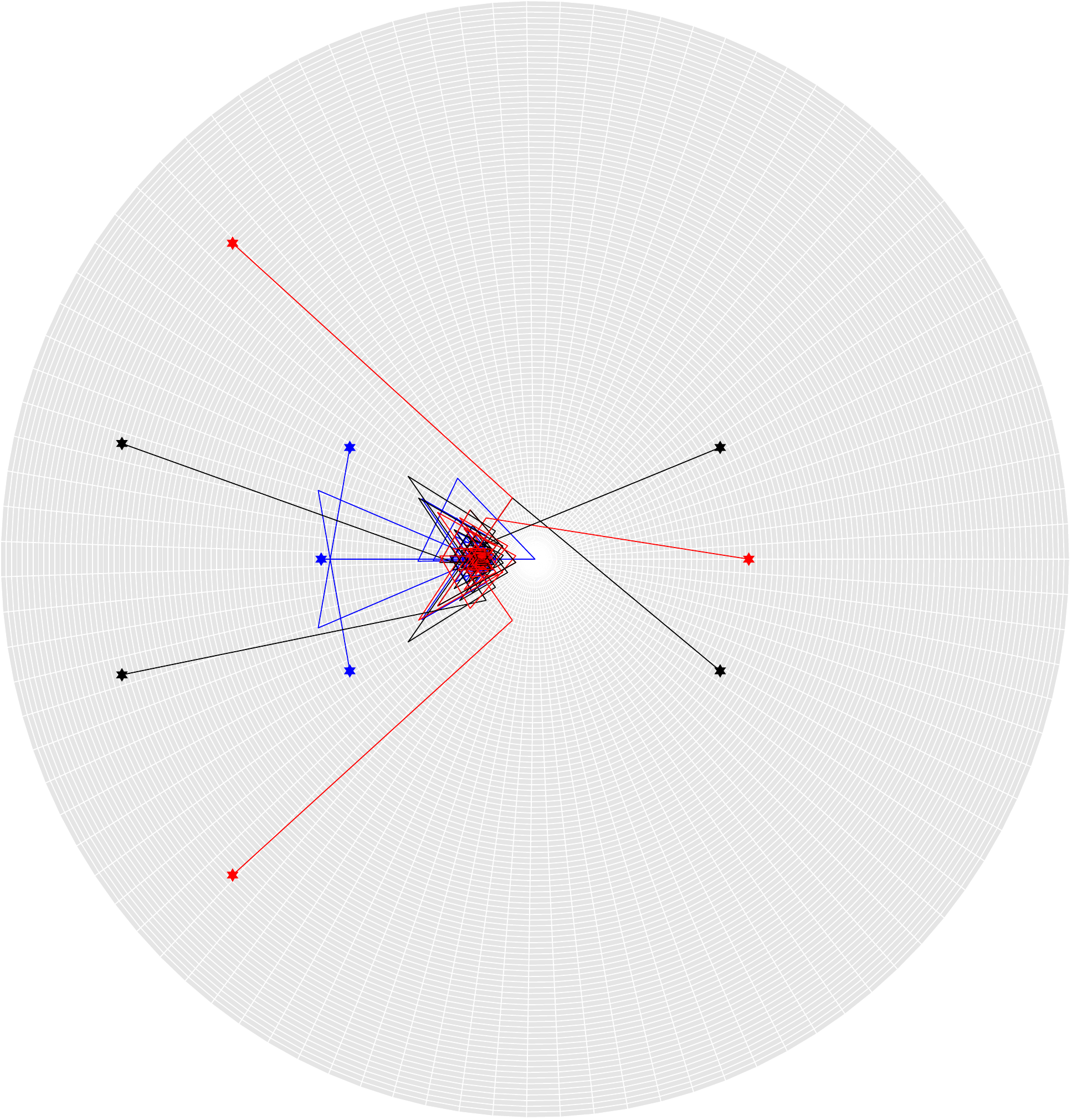}    
\caption{Active (red), inactive external (black), and internal data (blue) for an example with $d = 2$, $N = 10$. The IMR trajectories are only influenced by the external data, while the point of convergence itself is only dependent on the active data.} 
\label{fig:active}
\end{center}
\end{figure}

The properties of active and external data are observed even in very small data sets. If the right-hand point in Fig.~\ref{fig:cone_d2n3} is moved leftward, a sharp transition occurs beyond which the IMR is dependent only on the remaining two data points, and can be shown to occur at their geometric midrange. Fig.~\ref{fig:active} shows a more typical example, in which any or all of three input data internal to the data set (blue) may be removed without affecting the IMR convergence algorithm, and four external but non-active data points (black) may also be removed without affecting the IMR convergence point. In this example, there are only three active (red) data points.

The IMR algorithm may therefore be made more efficient by advance identification of all active points and restricting the max distance search to only target this subset. Data sets with extreme outliers will tend to have fewer active points, and the IMR would accordingly experience a dramatic speedup in such cases.

\section{Midrange Clustering on Matrix Data}

Clustering methodologies using nonlinear geometries related to eigenspectra of SPD matrices have shown promise. For example, see the adaptation of $K$-means clustering to several similarity measures considered by \cite{Stanitsas2017}, which include the affine-invariant Riemannian (AIRM) and log-Euclidean metrics. See also \cite{Nielsen2017} for an account of clustering
in Hilbert simplex geometry.  In this work, we consider $K$-means clustering with the Thompson distance $d_\infty$ as the similarity measure, and employ the IMR to calculate centroids for clusters. We further employ $X$-means and $K$-means++ algorithms with geometric midrange statistics and evaluate their performance.

The $K$-means clustering algorithm may be summarized as follows, where the inputs are a set of SPD data $\bar{Y}$ and a desired quantity of clusters $K$ (for a more thorough review, see \cite{Steinley2006}):

\begin{enumerate}
  \item Assign an initial cluster label $l_i \in \{1, ..., k\}$ to each $Y_i$. The set of points with label $L$ define a cluster $C_L \subset \bar{Y}$.
  \item For each cluster $C_L$, a centroid $\mu_L$ is calculated (here via IMR).
  \item For each data point $Y_i$, the nearest centroid $\mu_{L'}$ is identified, and the cluster label is reassigned $l_i \gets L'$.
  \item Repeat steps 2 and 3 until the cluster labels do not change between iterations.
\end{enumerate}

Two straightforward initialization methods are random assignment of labels, and random selection of $K$ data points as initial centroids, where labels are then assigned as in step 3. Both of these methods can have severe limitations due to false merging of clusters, where the severity decreases with the connectedness of the data set. In a highly disconnected data set, a true cluster which does not initially contain at least one centroid is unlikely to be accurately identified during the course of the algorithm. Oversplitting of clusters is a less severe but still significant source of error. Near-perfect accuracy can be achieved given advance knowledge of approximate cluster locations, so that one initial centroid can be placed in each; however this is not the usual case in applications, where the optimal value of $K$ is often not even known.

\subsection{$X$-means Clustering}

$X$-means is an iterative extension of $K$-means, first proposed by \cite{Pelleg2000}, which resolves the issue of erroneously merged clusters by repeated binary splits. In this work the authors make use of the Bayesian information criterion (BIC) to accept or reject these binary splits, as described in the following outline:

\begin{enumerate}
	\item Set $K$ to an initial lower-bound value $K_0$.
	\item Perform $K_0$-means clustering on $\bar{Y}$.
	\item For each cluster $C_L$, generate two initial split centroids and perform $K$-means with $K = 2$ on $C_L$ alone.
	\item If the BIC score for a split cluster exceeds that of the unsplit cluster, accept that split.
	\item Repeat steps 2-4 until no binary splits are accepted.
\end{enumerate}

In our $X$-means implementation, a low (empirical) limit is placed on the number of attempted splits that may be performed on a particular cluster. BIC scoring is modified to include a Thompson-based maximum likelihood estimator. We also split centroids following Thompson geometry, by randomly selecting one matrix on a fixed-radius $d_\infty$-sphere from the unsplit centroid, and taking this SPD and its antipodal point on the $d_\infty$-sphere as the initial split centroids before performing the intermediary $K = 2$ clustering. $d_\infty$-sphere sampling is performed by generating SPDs in a near-radially symmetric distribution around the identity. As the $d_\infty$ geodesic given by (\ref{Nussbaum}) is parameterized proportionally to Thompson distance, a retraction (or extension) along the geodesic is then made such that the generated point has the desired radius from the identity. By affine invariance of the metric, any point $\Sigma$ generated on the $d_\infty$ sphere may then be transported to the desired centroid $C$ by a congruence transformation $\Sigma \mapsto C^\frac{1}{2} \Sigma C^\frac{1}{2}$.

\begin{figure*}[h]
\begin{center}
\includegraphics[width=0.8\linewidth]{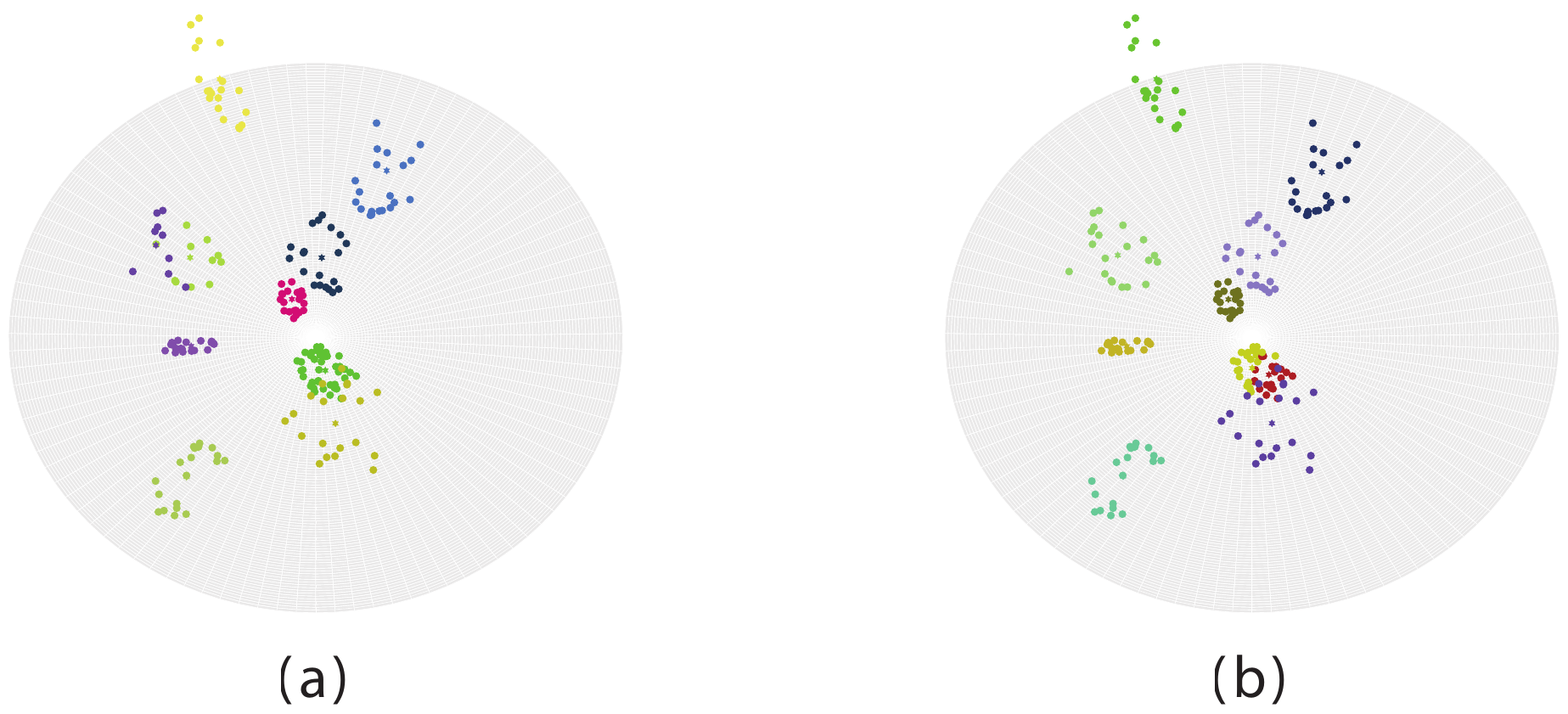}    
\caption{(a) $X$-means assigned labels for $10 
\times 20$ clustered SPDs, $d=2$.  (b) $K$-means++ assigned labels for $10 \times 20$ clustered SPDs, $d=2$.} 
\label{fig:xmeans_10x20}
\end{center}
\end{figure*}

Fig.~\ref{fig:xmeans_10x20} (a) shows representative $X$-means results for 200  $d=2$ SPD data in 10 disjoint clusters, visualized in the cone projection. These disjoint clusters are created by generating 10 cluster centers with the requirement that all pairwise Thompson distances are at least 1. For each cluster center, 20 data points are uniformly generated on a $d_\infty$-sphere of radius 0.2 around the cluster center. Although oversplitting can still occur, as exemplified by the center-left cluster in the plot, $X$-means massively reduces the risk of falsely merged clusters. Only one falsely merged cluster appears: the center-most cluster in this figure, where the bounding boxes of the two true clusters overlap.

For higher-dimensional SPDs, clustering results are best visualized in tabular form; for a selection of matrix dimensionalities of up to 20, $X$-means accuracy with $10\times20$ constructed data sets are summarized in Table~\ref{tb:xmeans}, averaged over 20 runs. In Table~\ref{tb:xmeans}, `points identified' refers to the number of points that were assigned to the correct cluster. `Clusters identified' refers to the number of clusters that were precisely identified in the sense that every point in the original assignment was correctly identified and no other points were identified erroneously. `Clusters lost' refers to the number of clusters that were not detected due to merging with nearby clusters. Due to volumetric scaling that naturally occurs with increased dimensionality, more accurate clustering is observed for high dimensions. Significant false-merging of clusters in the $d=2$ case appears to be attributable to poor performance of the BIC for small matrices, which gives a lower likelihood of splitting during step 4 of the algorithm.

\begin{table}[h]
\begin{center}
\caption{$X$-means clustering accuracy in high dimensions }\label{tb:xmeans}
\begin{tabular}{ccccc}
$d$ & 2 & 5 & 10 & 20 \\\hline
Points identified (/200)  & 109.2 & 172.0 & 170.0 & 200.0 \\
Clusters identified (/10)   & 4.3   & 8.2 & 8.3 & 10.0\\ 
Clusters lost (/10) 		& 4.5   & 1.4 & 1.5 & 0.0 \\\hline

\end{tabular}
\end{center}
\end{table}

\subsection{$K$-means++ Clustering}

A perhaps less invasive approach, rather than iterating the entire $K$-means procedure to reassign relatively few cluster labels, is to directly modify the initialization procedure which is the source of many oversplitting and false-merging errors. \cite{Arthur2007} achieve this by constructing a new initialization procedure from the ground up, which relies on no prior knowledge of clusters, other than the desired quantity of clusters. The resultant $K$-means++ initialization is as follows:

\begin{enumerate}
	\item Select an initial centroid $C_1$ at random from the data points.
	\item For each data point $Y_i$, assign a weight $w_i = (\text{distance to nearest centroid})^2$.
	\item Select another centroid among the data at random according to these (normalized) weights.
	\item Repeat steps 2-3 until $K$ centroids have been chosen.
\end{enumerate}

This procedure gives a vanishing likelihood that more than one initial centroid from the same cluster will be selected, and also can give such a strong initial guess at the clustering that $K$-means requires far fewer iterations to converge. In our implementation, the Thompson distance is employed for weight assignments.

Fig.~\ref{fig:xmeans_10x20} (b) shows a typical $K$-means++ result for the same 200 $d=2$ SPD data as in the above $X$-means example. No oversplitting or false merging occurs. The center-most pair of clusters, which were falsely merged in the $X$-means run, are now distinguished into two clusters in a qualitatively sensible way despite the overlap.

Clustering results are again obtained for a selection of matrix dimensionalities up to 100 and averaged over 20 runs. The $K$-means++ performance with 200 data points and 10 clusters is summarized in Table~\ref{tb:kmeanspp}.

\begin{table}[h]
\begin{center}
\caption{$K$-means++ clustering accuracy in high dimensions }\label{tb:kmeanspp}
\begin{tabular}{cccccc}
$d$ & 2 & 5 & 10 & 20 & 100 \\\hline
Points identified (/200) & 186.2 & 190.5 & 188.5 & 193.2 & 193.9 \\
Clusters identified (/10) &  8.5   & 8.9   & 8.8   & 9.3   & 9.3   \\ 
Clusters lost(/10) & 	 	 0.5   & 0.3   & 0.5   & 0.3   & 0.3   \\\hline

\end{tabular}
\end{center}
\end{table}

\medskip

\section{Conclusion}

We have introduced a novel inductive geometric midrange algorithm based on the Thompson geometry of the cone of positive definite matrices of a given dimension. The formulation is the natural generalization of the inductive characterization of the midrange of data points in linear spaces and has attractive computational properties. Important theoretical questions remain. For instance, how can one efficiently identify the subset of  active data from a given data set before implementing the IMR algorithm on the full set. Such an identification will result in dramatic improvements in the efficiency of the algorithm for large data sets, since it would generally remove the need for many unnecessary distance computations.

\bibliography{draft}             

\begin{thebibliography}{23}
\providecommand{\natexlab}[1]{#1}
\providecommand{\url}[1]{\texttt{#1}}
\providecommand{\urlprefix}{URL }
\expandafter\ifx\csname urlstyle\endcsname\relax
  \providecommand{\doi}[1]{doi:\discretionary{}{}{}#1}\else
  \providecommand{\doi}{doi:\discretionary{}{}{}\begingroup
  \urlstyle{rm}\Url}\fi

\bibitem[{{Arnaudon} et~al.(2013){Arnaudon}, {Barbaresco}, and
  {Yang}}]{Arnaudon2013}
{Arnaudon}, M., {Barbaresco}, F., and {Yang}, L. (2013).
\newblock Riemannian medians and means with applications to radar signal
  processing.
\newblock \emph{IEEE Journal of Selected Topics in Signal Processing}, 7(4),
  595--604.
\newblock \doi{10.1109/JSTSP.2013.2261798}.

\bibitem[{Arsigny et~al.(2006)Arsigny, Fillard, Pennec, and
  Ayache}]{Log-Euclidean2006}
Arsigny, V., Fillard, P., Pennec, X., and Ayache, N. (2006).
\newblock Log-{E}uclidean metrics for fast and simple calculus on diffusion
  tensors.
\newblock \emph{Magnetic Resonance in Medicine}, 56(2), 411--421.
\newblock \doi{10.1002/mrm.20965}.

\bibitem[{Arthur and Vassilvitskii(2007)}]{Arthur2007}
Arthur, D. and Vassilvitskii, S. (2007).
\newblock K-means++: The advantages of careful seeding.
\newblock In \emph{Proceedings of the Eighteenth Annual ACM-SIAM Symposium on
  Discrete Algorithms}, 1027--1035. Society for Industrial and Applied
  Mathematics, USA.

\bibitem[{Bhatia(2003)}]{Bhatia2003}
Bhatia, R. (2003).
\newblock On the exponential metric increasing property.
\newblock \emph{Linear Algebra and its Applications}, 375, 211 -- 220.

\bibitem[{Carroll and Chaturvedi(1998)}]{k-midranges}
Carroll, J.D. and Chaturvedi, A. (1998).
\newblock K-midranges clustering.
\newblock In A.~Rizzi, M.~Vichi, and H.H. Bock (eds.), \emph{Advances in Data
  Science and Classification}, 3--14. Springer Berlin Heidelberg, Berlin,
  Heidelberg.

\bibitem[{Dryden et~al.(2009)Dryden, Koloydenko, and Zhou}]{Dryden2009}
Dryden, I.L., Koloydenko, A., and Zhou, D. (2009).
\newblock Non-{E}uclidean statistics for covariance matrices, with applications
  to diffusion tensor imaging.
\newblock \emph{The Annals of Applied Statistics}, 3(3), 1102--1123.

\bibitem[{Golub and van~der Vorst(2000)}]{Golub2000}
Golub, G.H. and van~der Vorst, H.A. (2000).
\newblock Eigenvalue computation in the 20th century.
\newblock \emph{Journal of Computational and Applied Mathematics}, 123(1), 35
  -- 65.
\newblock \doi{https://doi.org/10.1016/S0377-0427(00)00413-1}.
\newblock Numerical Analysis 2000. Vol. III: Linear Algebra.

\bibitem[{Lemmens and Nussbaum(2012)}]{Lemmens2012}
Lemmens, B. and Nussbaum, R. (2012).
\newblock \emph{Nonlinear Perron-Frobenius Theory}.
\newblock Cambridge Tracts in Mathematics. Cambridge University Press.
\newblock \doi{10.1017/CBO9781139026079}.

\bibitem[{Lim(2013)}]{Lim2013}
Lim, Y. (2013).
\newblock Geometry of midpoint sets for {T}hompson's metric.
\newblock \emph{Linear Algebra and its Applications}, 439(1), 211 -- 227.
\newblock \doi{https://doi.org/10.1016/j.laa.2013.03.012}.

\bibitem[{Mishra and Sepulchre(2016)}]{Mishra2016}
Mishra, B. and Sepulchre, R. (2016).
\newblock Riemannian preconditioning.
\newblock \emph{SIAM Journal on Optimization}, 26(1), 635--660.
\newblock \doi{10.1137/140970860}.

\bibitem[{Moakher(2005)}]{Moakher2005}
Moakher, M. (2005).
\newblock A differential geometric approach to the geometric mean of symmetric
  positive-definite matrices.
\newblock \emph{SIAM J. Matrix Anal. Appl.}, 26(3), 735--747.
\newblock \doi{10.1137/S0895479803436937}.

\bibitem[{Mostajeran et~al.(2020)Mostajeran, Grussler, and
  Sepulchre}]{Mostajeran2020}
Mostajeran, C., Grussler, C., and Sepulchre, R. (2020).
\newblock Geometric matrix midranges.
\newblock \emph{SIAM Journal on Matrix Analysis and Applications}, 41(3),
  1347--1368.
\newblock \doi{10.1137/19M1273475}.

\bibitem[{Mostajeran and Sepulchre(2018)}]{Mostajeran2018}
Mostajeran, C. and Sepulchre, R. (2018).
\newblock Ordering positive definite matrices.
\newblock \emph{Information Geometry}, 1(2), 287--313.
\newblock \doi{10.1007/s41884-018-0003-7}.

\bibitem[{Nielsen and Sun(2017)}]{Nielsen2017}
Nielsen, F. and Sun, K. (2017).
\newblock Clustering in {H}ilbert simplex geometry.
\newblock \emph{arXiv preprint arXiv:1704.00454}.

\bibitem[{Nussbaum(1994)}]{Nussbaum1994}
Nussbaum, R.D. (1994).
\newblock Finsler structures for the part metric and {H}ilbert's projective
  metric and applications to ordinary differential equations.
\newblock \emph{Differential Integral Equations}, 7(5-6), 1649--1707.

\bibitem[{Pelleg and Moore(2000)}]{Pelleg2000}
Pelleg, D. and Moore, A. (2000).
\newblock X-means: Extending k-means with efficient estimation of the number of
  clusters.
\newblock In \emph{In Proceedings of the 17th International Conf. on Machine
  Learning}, 727--734. Morgan Kaufmann.

\bibitem[{Rao(2013)}]{Rao2013}
Rao, R.P.N. (2013).
\newblock \emph{Brain-Computer Interfacing: An Introduction}.
\newblock Cambridge University Press.
\newblock \doi{10.1017/CBO9781139032803}.

\bibitem[{{Stanitsas} et~al.(2017){Stanitsas}, {Cherian}, {Morellas}, and
  {Papanikolopoulos}}]{Stanitsas2017}
{Stanitsas}, P., {Cherian}, A., {Morellas}, V., and {Papanikolopoulos}, N.
  (2017).
\newblock Clustering positive definite matrices by learning information
  divergences.
\newblock In \emph{2017 IEEE International Conference on Computer Vision
  Workshops (ICCVW)}, 1304--1312.
\newblock \doi{10.1109/ICCVW.2017.155}.

\bibitem[{Steinley(2006)}]{Steinley2006}
Steinley, D. (2006).
\newblock K-means clustering: A half-century synthesis.
\newblock \emph{The British journal of mathematical and statistical
  psychology}, 59, 1--34.
\newblock \doi{10.1348/000711005X48266}.

\bibitem[{Stewart(2002)}]{Stewart2002}
Stewart, G.W. (2002).
\newblock A {K}rylov--{S}chur algorithm for large eigenproblems.
\newblock \emph{SIAM Journal on Matrix Analysis and Applications}, 23(3),
  601--614.
\newblock \doi{10.1137/S0895479800371529}.

\bibitem[{Stigler(2016)}]{stigler2016}
Stigler, S.M. (2016).
\newblock \emph{The seven pillars of statistical wisdom}.
\newblock Harvard University Press.

\bibitem[{Thompson(1963)}]{Thompson1963}
Thompson, A.C. (1963).
\newblock On certain contraction mappings in a partially ordered vector space.
\newblock \emph{Proceedings of the American Mathematical Society}, 14(3),
  438--443.

\bibitem[{{Zanini} et~al.(2018){Zanini}, {Congedo}, {Jutten}, {Said}, and
  {Berthoumieu}}]{Salem2018}
{Zanini}, P., {Congedo}, M., {Jutten}, C., {Said}, S., and {Berthoumieu}, Y.
  (2018).
\newblock Transfer learning: A {R}iemannian geometry framework with
  applications to brain-computer interfaces.
\newblock \emph{IEEE Transactions on Biomedical Engineering}, 65(5),
  1107--1116.

\end{thebibliography}

\appendix

\end{document}